\documentclass{article}
\usepackage[a4paper, total={7in, 10in}]{geometry}
\usepackage{xcolor}
\usepackage{xspace}
\usepackage[mathscr]{eucal}
\usepackage{amsmath}
\usepackage{amsthm}
\usepackage{graphicx}
\usepackage{mathtools}
\usepackage{researchpack}
\usepackage{subcaption}
\usepackage{url} 
\usepackage{algcompatible}
\usepackage{algorithm}
\usepackage[super]{nth}
\usepackage{lineno}
\usepackage{pbox}
\usepackage[norule, flushmargin]{footmisc}

\newcommand{\caipi}{\textsc{caipi}\xspace}
\newcommand{\SelectQuery}{\ensuremath{\textsc{\textcolor{violet}{SelectQuery}}}}
\newcommand{\Fit}{\ensuremath{\textsc{\textcolor{violet}{Fit}}}}
\newcommand{\Explain}{\ensuremath{\textsc{\textcolor{blue}{Explain}}}}
\newcommand{\ToCounterExamples}{\ensuremath{\textsc{\textcolor{orange}{ToCounterExamples}}}}
\newcommand{\ToBinaryCorrectionMask}{\ensuremath{\textsc{\textcolor{orange}{ToBinaryCorrectionMask}}}}

\newcommand{\lime}{\textsc{lime}\xspace}
\newcommand{\ce}{\textsc{ce}\xspace}
\newcommand{\rrr}{\textsc{rrr}\xspace}
\newcommand{\hint}{\textsc{hint}\xspace}

\newcommand{\gradcam}{\textsc{grad-Cam}\xspace}
\newcommand{\gradcams}{\textsc{grad-Cams}\xspace}

\newcommand{\rebuttal}[1]{{#1}}

\newcommand\blfootnote[1]{%
  \begingroup
  \renewcommand\thefootnote{}\footnote{#1}%
  \addtocounter{footnote}{-1}%
  \endgroup
}

\title{Making deep neural networks right for the right scientific reasons by interacting with their explanations}

\date{}

\author{\textbf{Patrick Schramowski$^{*}$, Wolfgang Stammer$^{*}$, Stefano Teso, Anna Brugger,} \\ \textbf{Franziska Herbert, Xiaoting Shao, Hans-Georg Luigs,} \\ \textbf{Anne-Katrin Mahlein \& Kristian Kersting}
}

\begin{document}

\maketitle

\begin{abstract}
Deep neural networks have shown excellent performances in many real-world applications. Unfortunately, they may show ``Clever Hans''-like behavior---making use of confounding factors within datasets---to achieve high performance. In this work, we introduce the novel learning setting of ``explanatory interactive learning'' (XIL) and illustrate its benefits on a plant phenotyping research task. XIL adds the scientist into the training loop such that she interactively revises the original model via providing feedback on its explanations. Our experimental results demonstrate that XIL can help avoiding Clever Hans moments in machine learning and encourages (or discourages, if appropriate) trust into the underlying model.
\end{abstract}

Imagine a plant phenotyping team attempting to characterize crop resistance to plant pathogens. The plant physiologist records a \rebuttal{large} amount of hyperspectral imaging data. Impressed by the results of deep learning in other scientific areas, she wants to establish similar results for phenotyping. Consequently, she asks a machine learning expert to apply deep learning to analyze the data. Luckily, the resulting predictive accuracy is very high. The plant physiologist, however, remains skeptical. The results are ``too good, to be true''. Checking the decision process of the deep model using explainable artificial intelligence (AI), the machine learning expert is flabbergasted to find that the learned deep model uses clues within the data that do not relate to the biological problem at hand, so-called confounding factors. The physiologist loses trust in AI and turns away from it, proclaiming it to be useless.%
\blfootnote{\textit{Preprint. Work in progress.} \\ \textcolor{black}{$^*$Equal contribution.}}

This example encapsulates an important issue of current explainable AI \rebuttal{\cite{guidotti2018survey,gilpin2018explaining}}. Indeed, the seminal paper of Lapuschkin \textit{et al.}~\cite{lapuschkin2019unmasking} helps in ``unmasking Clever Hans predictors and assessing what machines really learn''.
However, rather than proclaiming, as the plant physiologist might, that the machines have learned the right predictions for the wrong reasons and can therefore not be trusted, we here showcase that interactions between the learning system and the human user can correct the model, towards making the right predictions for the right reasons \rebuttal{\cite{ross2017right}}. This may also increase the trust in machine learning models. 
Actually, trust lies at the foundation of major theories of interpersonal relationships in psychology \cite{simpson2007psychological, hoffman2013trust} and we argue that interaction and understandability are central to trust in learning machines.
Surprisingly, the link between interacting, explaining and building trust has been largely ignored by the machine learning literature. Existing approaches focus on passive learning only and do not consider the interaction between the user and the learner~\cite{Bucilua2006Model,ribeiro2016should,lundberg2016unexpected}, whereas, interactive learning frameworks such as active~\cite{settles2011closing} and coactive learning~\cite{shivaswamy2015coactive}
do not consider the issue of trust. In active learning, for instance, the model presents unlabeled instances to a user, and in exchange obtains their label. This is completely opaque---the user is oblivious to the model’s beliefs and reasons for predictions and to how they change in time, and cannot see the consequences of her instructions. In coactive learning, the user sees and corrects the system’s prediction, if necessary, but the predictions are not explained to her. So, why should users trust models learned interactively?

Furthermore, although an increasing amount of research investigates methods for explaining machine learning models, even here the notion of interaction has been largely ignored. Reconsider the study by Lapuschkin \textit{et al.}~\cite{lapuschkin2019unmasking}. They showed that one can find ``Clever Hans''-like behavior in popular computer vision models basing their decisions on confounding factors. %These factors may act as good indicators within the particular dataset but would prove to be useless in real-world settings. 
Based on these findings, the authors recommended a word of caution towards the interest in such models, but they did not offer a solution for correcting their behavior. Particularly in real-world applications, where monitoring for every possible confounding factor or acquiring a new dataset due to existing confounders is time and resource consuming, it is inevitable to move beyond revealing the (wrong) reasons \rebuttal{by making a} step towards correcting the reasons underlying a models decisions.
%\todo{Cite HINT \cite{selvaraju2019taking} and RRR \cite{ross2017right} and \cite{erion2019learning} --> currently cited \cite{selvaraju2019taking} and \cite{erion2019learning} in discussion of limitations}
%%
\begin{figure}[t!]
\centering
\includegraphics[width=1.\linewidth]{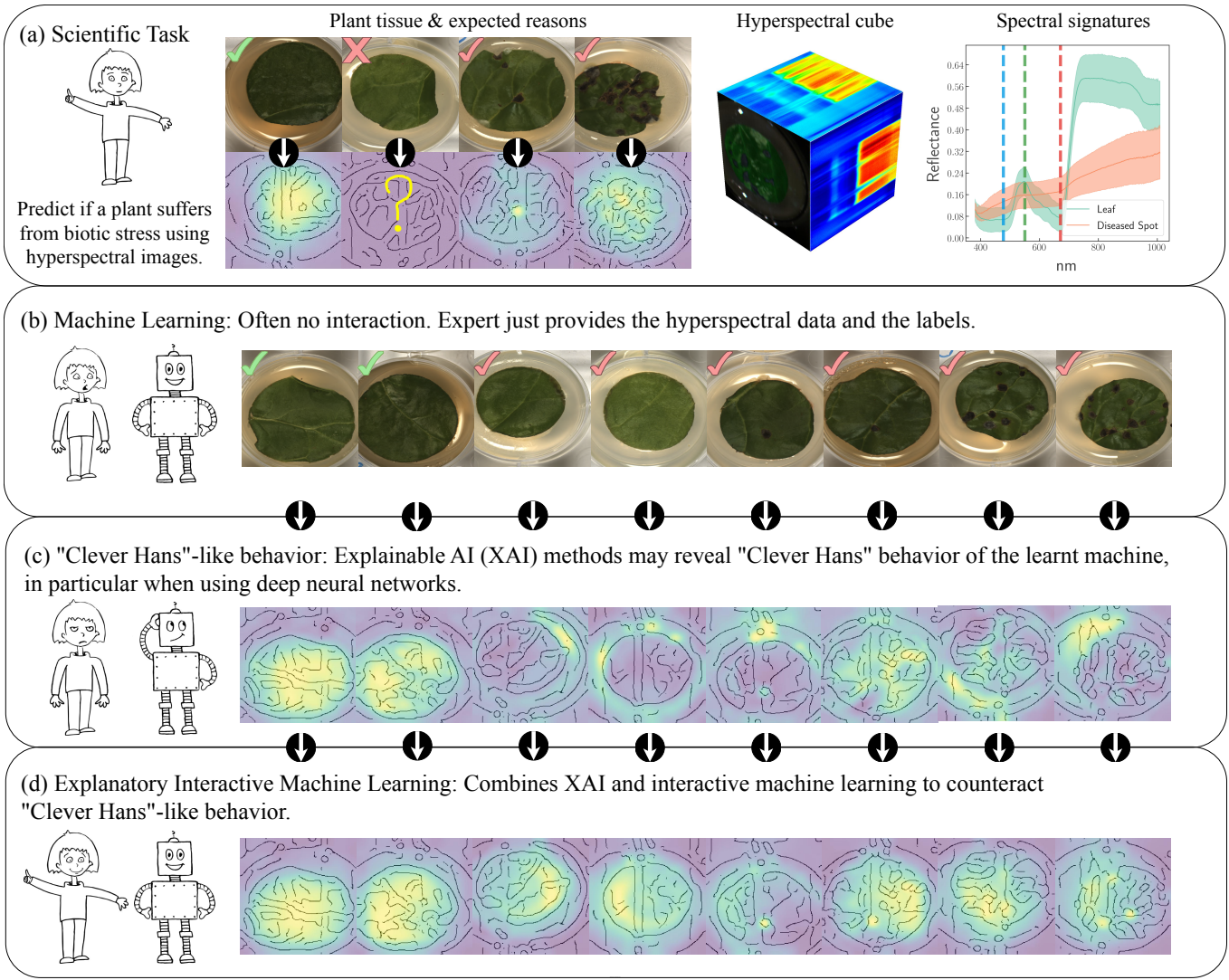}
%\includegraphics[width=1.\linewidth]{images/cams/wrongAndRightReasonsHS.pdf}
%alle vom mittwoch: Z18,4,1,1   Z17,1,0,0   Z16,2,1,1   Z15,2,1,2   Z8,4,0,0   Z8,4,1,2   Z1,3,1,1   Z2,1,0,2 
\caption{\textbf{Explanatory Interactive Learning (XIL)}---Human users revise learning machines towards trustworthy decision strategies. \rebuttal{The cartoons sketch the main unerlying idea for each case (row)}. \textbf{(a-left)} Data samples,
\rebuttal{expert-classifications (checks and Xs with colors indicating the class)} and explanations \rebuttal{(overlaid with an edge filtered original image for better interpretability)} that an expert expects of an ML model. \rebuttal{Yellow corresponds to relevant regions, blue to irrelevant regions for a classification.} Not even an expert can be \rebuttal{certain} about \rebuttal{potential samples from a early disease stage} and \rebuttal{what a valid explanation should be}. \textbf{(a-middle)} \rebuttal{Illustration} of hyperspectral data \rebuttal{consisting of} spatial and spectral dimensions. The planes on the top and left sides of the cube correspond to slices taken from the center of the cube but placed on the edges for visualization. \textbf{(a-right)} The characteristic reflectance of healthy tissue vs.~disease spots. The vertical red, green and blue lines depict the three wavelengths of the RGB dataset. \textbf{(b,c)} Classifications of a deep neural network \rebuttal{(b)} and its explanations \rebuttal{\textbf{(c)}}. The learned model clearly uses confounding factors\rebuttal{, identified as the embedding agar solution,} to explain its decision. \textbf{(d)} The human user provides feedback on the reasons. In turn, the machine gets new information and can continue learning.  The human-revised deep network yields classifications matching \rebuttal{a biologically plausible} strategies. (\rebuttal{All shown RGB images correspond to real RGB images, while} the edge overlays resulted from pseudo-RGB images generated from the original \rebuttal{hyperspectral} dataset, \textit{cf.} Methods  RGB/HS classification.)\label{fig:story}}
\end{figure}

\rebuttal{Doing so} is exactly the main technical contribution of the present study. We introduce the novel learning setting of ``explanatory interactive learning'' (XIL) and illustrate its benefits in an important scientific endeavor, namely, plant phenotyping. Starting from a learning system that does not deliver biologically plausible explanations for a relevant, real-world task in plant phenotyping, we add the scientist into the training loop, who interactively revises the original model \rebuttal{by interacting} via \rebuttal{it's} explanations so that it produces trustworthy decisions without a major drop in performance. 
Specifically, \rebuttal{XIL} takes the form illustrated in Fig.~\ref{fig:story}. In each step, the learner explains its interactive query to the domain expert, and she responds by correcting the explanations, if necessary, to provide feedback. This allows the user not only to check whether the model is right or wrong on the chosen instance but also if the answer is right (or wrong) for the wrong reasons, e.g., when there are ambiguities in the data such as confounders ~\cite{ross2017right}. By witnessing the evolution of the explanations, similar to a teacher supervising the progress of a student, the human user can see whether the model eventually ``gets it''. The user may even correct the explanation presented to guide the learner. This correction step is crucial for more directly affecting the learner's beliefs and is integral to modulating trust~\cite{hoffman2013trust,kulesza2015principles}. 

\rebuttal{Specifically, we make the following contributions:
%\begin{
     %\item
     (i) Introduction of XIL with counterexamples (CE) to revise ``Clever Hans'' behavior in a model-agnostic fashion.
    %\item 
    (ii) Adaption of the ``right for the right reasons'' (RRR) loss to latent layers of deep neural networks.
    %\item 
    (iii) Showcasing XIL on the computer vision benchmark datasets PASCAL VOC 2007 \cite{pascal-voc-2007} and MSCOCO 2014 \cite{LinMBHPRDZ14}.
    (iv) Evaluation of XIL on a highly relevant dataset for plant phenotyping, demonstrating its potential as an enabler of scientific discovery.
    %\item 
    (v) Gathering of the plant phenotyping dataset and the creation of a version with confounders.
    %\item
    (vi) A user study on trust development within XIL \cite{herbert2019}. 
    %\item 
%    (vii) A discussion of the limitations of XIL and future perspectives.
%\end{itemize}
}

A preliminary version of this manuscript has been published as a conference paper
\cite{teso2019explanatory}. The present paper significantly extends the conference version by (ii-v)
%demonstrating for the first time that deep networks can have 
%``Clever Hans''-like moments in plant phenotyping, extending XIL towards latent layers of deep neural networks, and %showing how to ``un-Hans'' deep neural networks for plant phenotyping \rebuttal{and on two ML benchmark datasets} via %interacting with their explanations. 
Moreover, the \rebuttal{ad-hoc} XIL user study \rebuttal{presented in \cite{teso2019explanatory} was completely re-designed, newly conducted, and now includes a thorough statistical analysis (vi)}. \rebuttal{To encourage further research, we provide the created plant phenotyping dataset.}
%\todo{a little redundant to previous paragraph?}

We proceed as follows. We start by formally introducing Explanatory Interactive Machine Learning (XIL) and instantiate it in the \caipi method \cite{teso2019explanatory}
as well as the \rrr method \cite{ross2017right}.
After introducing XIL, we discuss quantitative results on test datasets, before providing details on how domain experts can revise learning machines and in turn enable the machines to correct their abilities to solve the scientific real-world task of plant disease prediction. Finally, we demonstrate the importance of explaining decisions for building trustful machines via a user study. 
Our contributions thus address a main part of building trustworthy AI methods by providing an end-to-end, interactive method to evaluate and revise \rebuttal{machine learning} models. 
%\todo{Change wording to emphasize that XIL still necessary for IAI} 
This provides an important \rebuttal{add-on} to Rudin's \cite{rudin2019} message ``\textit{Stop explaining black-box machine learning models for high stakes decisions and use interpretable models instead}'', namely
\rebuttal{\textit{``Start interacting with explanations of machine learning models to avoid `Clever Hans'-like behavior.''}}
%\begin{quote}\textit{Continue to explain black-box models since they can alleviate ``Clever Hans''-like problems when used to revise the model interactively.}
%\end{quote}

\subsection*{Explanatory Interactive Machine Learning (XIL)}
In XIL, a learner can interactively query the user (or some other information source) to obtain the desired outputs of the data points. The interaction takes the following form. At each step, the learner considers a data point (labeled or unlabeled), predicts a label, and provides explanations of its prediction.  The user responds by correcting the learner if necessary, providing a slightly improved---but not necessarily optimal---feedback to the learner.

Let us now instantiate this schema to {\it explanatory active
learning}---combining active learning with local explainers (\textit{cf.} Methods). Indeed,
other interactive learning can be made explanatory too, including coactive learning~\cite{shivaswamy2015coactive}, active imitation
learning~\cite{judah2012active}, and mixed-initiative interactive
learning~\cite{cakmak2011mixed}, but this is beyond the scope of this
paper.

\paragraph{Explanatory Active Learning.}
In Explanatory Active Learning, we require black-box access to an active learner and an explainer. We assume that the active learner provides a procedure $\SelectQuery(f, \calU)$ for selecting an informative instance $x \in \calU$ based on the current model $f$, and a procedure $\Fit(\calL)$ for fitting a new model (or update the current model) on the examples in $\calL$.  The explainer is assumed to provide a procedure $\Explain(f, x, \hat{y})$ for explaining a particular prediction $\hat{y} = f(x)$.  The framework is intended to work for
any reasonable learner and explainer.  

When using  \lime{} %(described above)
for computing an interpretable model locally around the queries to visualize explanations for current predictions, this results in \caipi as summarized in Alg.~\ref{alg:framework}.
\begin{algorithm}[t]
    \begin{algorithmic}[1]
            \State $f \gets \Fit(\calL)$
            \REPEAT
                \STATE $x \gets \SelectQuery(f, \calU)$
                \STATE $\hat{y} \gets f(x)$
                \STATE $\hat{z} \gets \Explain(f, x, \hat{y})$
                \STATE Present $x$, $\hat{y}$, and $\hat{z}$ to the user
                \STATE Obtain $\bar{y}$ and explanation correction $\calC$
                %\STATE if \caipi: $\{(\bar{x}_i, \bar{y})\}_{i=1}^c \gets \ToCounterExamples(\calC)$
                %\STATEx \hspace{\algorithmicindent}else if \rrr: $\{(x, \bar{y}, A)\} \gets \ToBinaryCorrectionMask(\calC)$
                %\STATE if \caipi: $\calL \gets \calL \cup \{(x, \bar{y})\} \cup \{(\bar{x}_i, \bar{y})\}_{i=1}^c$ 
                %\STATEx \hspace{\algorithmicindent}else if \rrr: $\calL \gets \calL  \cup \{(x, \bar{y}, A)\} $
                \STATE if \textcolor{red}{\caipi}: 
                    \STATE \hspace{\algorithmicindent} $\{(\bar{x}_i, \bar{y})\}_{i=1}^c \gets \ToCounterExamples(\calC)$
                    \STATE \hspace{\algorithmicindent} $\calL \gets \calL \cup \{(x, \bar{y})\} \cup \{(\bar{x}_i, \bar{y})\}_{i=1}^c$
                \STATE else if \textcolor{red}{\rrr}: 
                    \STATE \hspace{\algorithmicindent} $\{(x, \bar{y}, A)\} \gets \ToBinaryCorrectionMask(\calC)$
                    \STATE \hspace{\algorithmicindent} $\calL \gets \calL  \cup \{(x, \bar{y}, A)\} $
                \STATE $\calU \gets \calU \setminus \{x\}$ \label{eq:updatesets}
                \STATE $f \gets \Fit(\calL)$
            \UNTIL{budget $T$ is exhausted or $f$ is good enough}
            \State \textbf{return} $f$
 %       \EndProcedure
    \end{algorithmic}
    \caption{\label{alg:framework} \caipi\ takes as input a
    set of labeled examples $\calL$, a set of unlabeled instances $\calU$,
    and iteration budget~$T$.}
\end{algorithm}
At each iteration $t = 1, \ldots, T$ an instance $x \in \calU$ is chosen using the query selection strategy implemented by the \SelectQuery\ procedure. Then its label $\hat{y}$ is predicted using the current model $f$, and \Explain\ is used to produce an explanation $\hat{z}$ of the prediction. The triple $(x, \hat{y}, \hat{z})$ is presented to the user as a (visual) artifact. The user checks the prediction and the explanation for correctness and provides the required feedback. Upon receiving the feedback, the system updates $\calU$ and $\calL$ accordingly and re-fits the model. The loop terminates when the iteration budget $T$ is reached or the model is good enough.

During interactions between the system and the user, three cases can occur: 
{\bf (1) Right for the right reasons:} The prediction and the explanation are both correct. No feedback is requested. {\bf (2) Wrong for the wrong reasons:} The prediction is wrong. As in active learning, we ask the user to provide the correct label. \rebuttal{While the explanation may provide some signal as to why the prediction was wrong, we currently do not require the user to act on it---this is an interesting avenue for future work---but treat the explanation to be simply wrong.} {\bf (3) Right for the wrong reasons:}  The prediction is correct but the explanation is wrong---the main target of XIL. 

\paragraph{\rebuttal{Model-agnostic} XIL using counterexamples (CE).}
The ``right for the wrong reasons'' case is novel in active learning, and we propose {\it explanation corrections} to deal with it. They can assume different meanings depending on whether the focus is on component relevance, polarity, or relative importance (ranking), among others. In our experiments we ask the annotator to indicate the components that have been wrongly identified by the explanation as relevant, that is,
\begin{linenomath*}
\begin{equation}
    \calC = \{j : |w_j| > 0 \land \text{the user believes the $j$th component to be irrelevant}\} \ . \nonumber
\end{equation}
\end{linenomath*}

Given the correction $\calC$, we are faced with the problem of explaining it back to the learner. We propose a simple strategy to achieve this. This strategy is embodied by \ToCounterExamples. It converts $\calC$ to a set of \emph{counterexamples} that teach the learner not to depend on the irrelevant components. In particular, for every $j \in \calC$ we generate $c$ examples $(\bar{x}_1, \bar{y}_1), \ldots, (\bar{x}_c, \bar{y}_c)$, where $c$ is an application-specific constant. Here, the labels $\bar{y}_i$ are identical to the prediction $\hat{y}$. The instances $\bar{x}_i$, $i = 1, \ldots, c$ are also identical to the query $x$, except that the $j$th component (i.e. $\psi_j(x)$) has been either randomized, changed to an alternative value, or substituted with the value of the $j$th component appearing in other training examples of the same class.  
This counterexample strategy (\ce) produces $c \cdot|\calC|$ counterexamples, which are added to $\calL$, as summarized in Alg.~\ref{alg:framework}. Importantly, this method is model-agnostic and can be used also when applying a non-differentiable model.

\paragraph{XIL \rebuttal{using gradients.}} % the ``Right for the Right Reason'' (RRR) loss.}
\rebuttal{If the model is differentiable, the learner can also be regularized to be right for the right reasons using the ``Right for the Right Reasons''} loss (\rrr) introduced by Ross \textit{et al.}~\cite{ross2017right}. %\rrr, in contrast to \ce, can be used when the model is differentiable.
%This additional regularization term 
\rebuttal{Here one} adds a penalty to gradients that lie outside of a binary mask that indicates which features of the input are relevant. We modified the original loss function to:
\begin{linenomath*}
\begin{align}
\label{eqn:rrr}
L(\theta,\ X,\ y,\ A) =& \underbrace{\sum_{n=1}^{N} \sum_{k=1}^{K} -c_ky_{nk} \log({\hat{y}}_{nk})}_\text{Right answers}
\quad + \quad \underbrace{\lambda_1 \sum_{n=1}^{N} \sum_{d=1}^{D} \left(A_{nd} \frac{\delta}{\delta h_{nd}} \sum_{k=1}^{K} c_k\log({\hat{y}}_{nk})\right)^2}_\text{Right reasons}
+ \underbrace{\lambda_2 \sum_i \theta^2_i}_\text{Weight regularization}\ , 
\end{align}
\end{linenomath*}
where $\theta$ describes the parameters of the network, $X$ the input, $y$ the ground truth and $A$ the binary mask used in the regularization term that discourages the input gradient from being large in regions marked by $A$. Instead of regularizing the gradients with respect to $X$, as originally described in \cite{ross2017right}, we regularize the gradients of the final convolutional layer $h$, corresponding to Gradient weighted Class Activation Maps (\gradcam) (\cite{selvaraju2017grad}, \textit{cf.} Methods). Further $c$ is a rescaling weight given to each class of the unbalanced dataset and $\hat{y}$ corresponds to the network prediction. The objective function is split into three terms. The first and the last are the familiar cross-entropy and weight ($\theta$) regularization terms. The second term is the new regularization term. The $\lambda$ values are used to weight the different regularizations. Ross \textit{et al.}~\cite{ross2017right} state that the regularization parameter $\lambda_1$ should be
set such that the ``right answers'' and ``right reasons'' terms have similar orders of magnitude. 

\rebuttal{\rrr can easily be incorporated into XIL (see again Alg.~\ref{alg:framework}), and, as demonstrated by Selvaraju {\it et al.}'s \hint approach~\cite{selvaraju2019taking}, Eq.~\ref{eqn:rrr} can be extended if the user is confident about how a (visual) explanation should look like. See Methods for an extended discussion of related work.} %In this case, an additional regularization term may help to get the gradient-based explanations of the network close to the human-provided ones.}
%\rebuttal{To summarize, the XIL framework with \ce is learner-agnostic and achieves model improvements via data augmentation. In contrast, XIL with \rrr can only be applied to differentiable models and achieves model improvements by acting directly on model internals. In both cases, XIL assumes that the user feedback and the explanation method for the model’s decisions are faithful. The ``wrong for the right reasons'' case is not exploited, and we mainly investigate XIL for high accuracy scenarios and datasets with confounding factors in our experiments. Investigating XIL when handling this case constitutes an interesting avenue for future work.}

%Let us now present several showcases that demonstrate the effectiveness of explanatory machine learning methods for understanding, validating, and correcting the behavior of a learned model. %We finally investigate how providing explanations to users changes their trust in the model's choices.
%\todo{add remarks to HINT and suppl. material}

\paragraph{\rebuttal{\bf Demonstrating XIL on Computer Vision datasets.}}
We begin by considering simulated users---as it is common for active learning---to evaluate the contribution of explanation feedback.
Indeed, counterexample strategies (e.g. \caipi) can trivially accommodate more advanced models than the one employed here. %As is common in active learning, 
We simulate a human annotator that provides correct labels.  Explanation corrections are also assumed to be correct and complete (i.e. they identify all false-positive components), for simplicity.%\footnote{In practice corrections may be incomplete or noisy, especially when dealing with non-experts. This can be handled by, e.g., down-weighting the counterexamples.}.

\begin{table}[t!]
%\small
\setlength\tabcolsep{3pt}
\centering
    %\begin{tabular}{r|c|c|c}
    %     \multicolumn{4}{c}{\textbf{(a) User Study}} \\
    %     \multicolumn{4}{c}{} \\
    %     & \textbf{Q1} & \textbf{Q2}   & \textbf{Q3} \\
    %     \hline
    %     \textbf{TRC1}  & 65\%      & 35\%        & 82\% \\
    %     \hline
    %     \textbf{TRC2}  & 29\%      & 12\%        & 41\% \\
    %     \hline
    %     \textbf{TRC3}  & 77\%      & 65\%        & 71\% \\
    %     \hline
    %     \textbf{TRC4}  & 77\%      & 65\%        & 71\% \\
    % \end{tabular} \quad \quad
    \begin{tabular}{l|c|ccc|c}
        \multicolumn{6}{c}{{(a) Fashion-MNIST (Toy) Dataset}} \\
        \multicolumn{6}{c}{} \\
        \multicolumn{6}{c}{} \\
        &  no &\multicolumn{3}{|c|}{Counterexamples} & \rrr\\
        &  corr. & $c=1$ & $c=3$  & $c=5$ & IG\\
        \hline
        Train   & {\bf 97}\%     & 93\% &  92\% &  92\% & 89\% \\
        Test    & 48\%     & 82\% & {\bf 85}\% &  {\bf 85}\% & {\bf 85}\% %\\
    \end{tabular} \quad \quad
    \begin{tabular}{l|c|c}
        \multicolumn{3}{c}{{(b) Scientific Dataset}} \\
        \multicolumn{3}{c}{} \\
        \multicolumn{3}{c}{} \\
        & no. & \rrr \\
        & corr. & \gradcam \\ \hline
        RGB & {\bf 89}\% & 87\%* \\
        HS & {\bf 99}\% & 95\%
        % %CNN RGB: (88,39+88,93+92,93+89,29+85,44)/5
        % %CNN RGB RRR: (81,29 + 86,72 + 93,20 + 85,48 + 89,05)/5 (epochs: (70, 90, 100, 60, 200))
        % %CNN RGB RRR old: (90,51+87,93+88,99+87,51+87,45)/5
        % %CNN HS: (99,62+99,26+98,45+98,16+98,28)/5
        % %CNN HS RRR: (92,59+98,84+96,12+92,69+94,92)/5
        % %CNN HS {Masked-Masked: (91,26+87,50+61,24+67,20+86,59)/5 == 78.76
        % %CNN HS Masked-Default: (73,79+78,77+73,38+81,99+81,28)/5 == 77.84

    \end{tabular}\quad \quad
    \rebuttal{
    \begin{tabular}{l|c|c}
        \multicolumn{3}{c}{{(c) HS Scientific Dataset}} \\
        \multicolumn{3}{c}{{non-confounded test set}} \\
        \multicolumn{3}{c}{} \\
        per-channel &  no & \rrr\\
        average &  corr. & \gradcam\\
        \hline
        \pbox{15cm}{non-tissue}  & 81\%     & {\bf 87}\% \\ % & 81.08\% & 87.19\%
        full image    & 50\%     & {\bf 82}\%  % & 50.40\% & 82.96\%
    \end{tabular}
    }
    \caption{Explanatory feedback can boost trust and performance. Highest performances are bold.  %(a)  User study: percentage of ``yes'' answers. 
    (a) Accuracy on the fashion MNIST dataset of an
    MLP without corrections (no corr.), with our (\ce) using
    varying $c$ (middle), and \rrr with input gradient (IG) constraints~\cite{ross2017right}. 
    (b) The \rebuttal{average} model balanced accuracy of applying \rrr with \gradcam over five cross-validation runs. With ``*'' we denote situations where decisions made based on the background could not be fully removed. %, only those cross-validations with satisfactory explanations were considered.
    \rebuttal{(c) The average model balanced accuracy over five cross-validation runs on a non-confounded test set of the hyperspectral (HS) scientific data. The confounding background features were set to either the per-channel average of the non-tissue regions or the full image of the training samples. The accuracies are reported for HS-CNN.} 
     \label{tab:results}}
\end{table}

\begin{figure}[t!]
    %\centering
    %\includegraphics[width=.6\linewidth]{images/pascal_voc/pascal_voc_examples.pdf}

     \centering
     \begin{subfigure}{0.25\textwidth}
         \centering
         { (a) Test Image}\\
         \includegraphics[width=.9\linewidth]{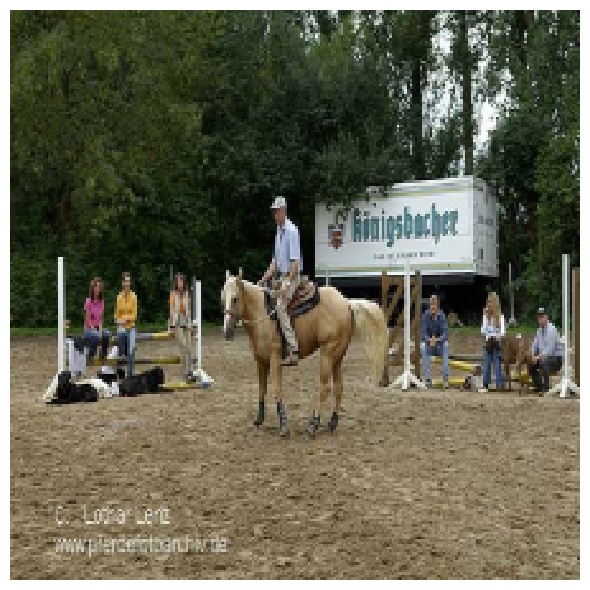}
         \includegraphics[width=.9\linewidth]{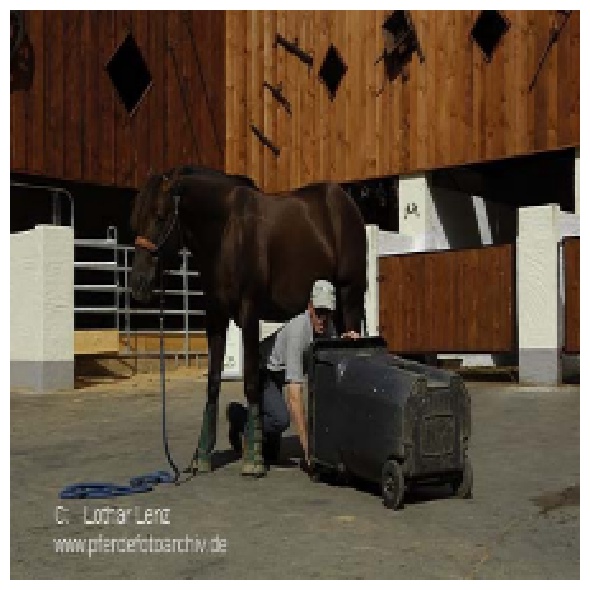}
     \end{subfigure}
     %\quad    \quad
     \begin{subfigure}{0.25\textwidth}
         \centering
         {(b) w/o Feedback}\\
         \includegraphics[width=.9\linewidth]{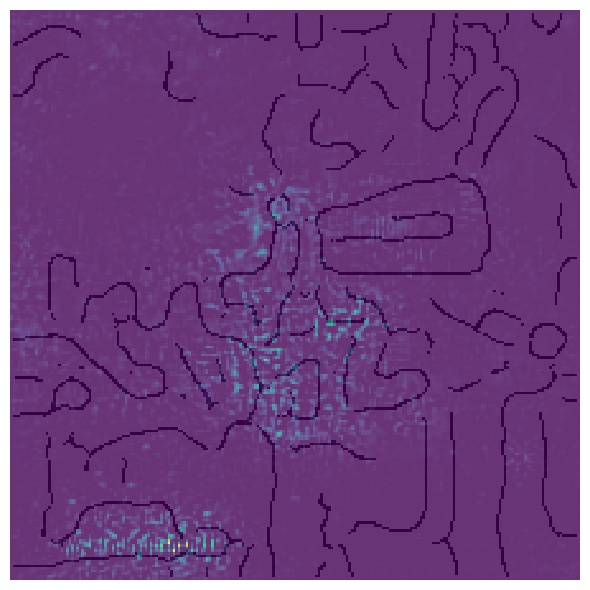}
         \includegraphics[width=.9\linewidth]{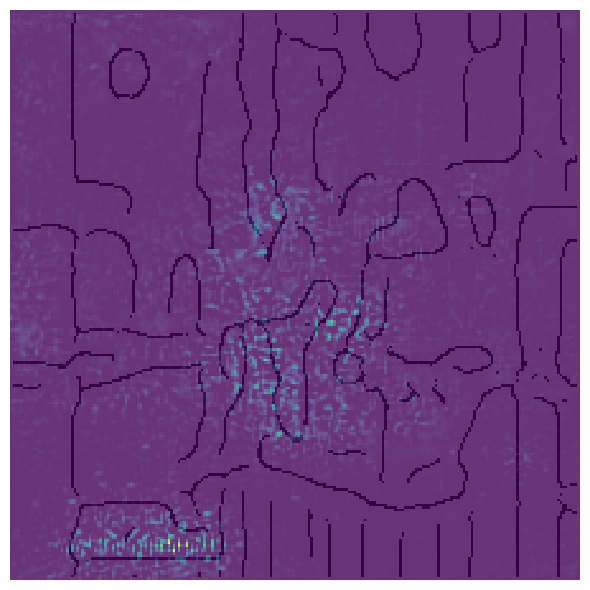}
     \end{subfigure}
     %\quad    \quad
    \begin{subfigure}{0.25\textwidth}
         \centering
         {(c) XIL}\\
         \includegraphics[width=.9\linewidth]{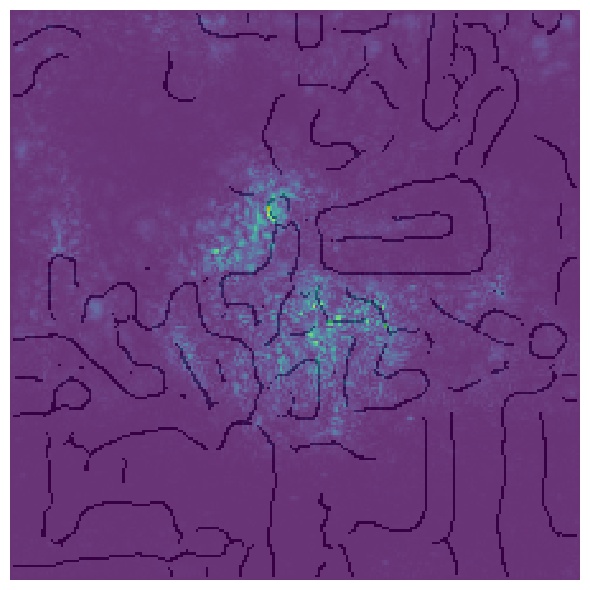}
         \includegraphics[width=.9\linewidth]{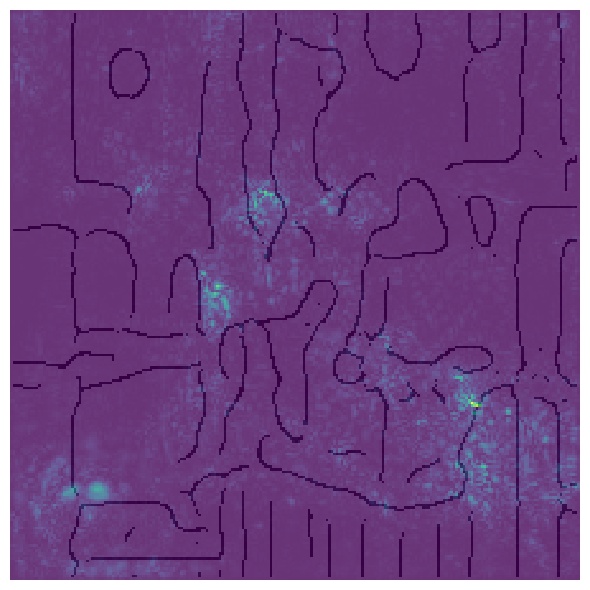}
     \end{subfigure}
    %(a) \hspace (b) \hspace (c)\hspace\\
    \caption{\rebuttal{XIL helps avoiding Clever Hans moments on unseen PASCAL VOC images (a). Ignoring user feedback, the model focuses on a source tag present in the lower left corner (b). Training it via interacting with its explanations, it does not consider the source tag to be relevant anymore (c). The visual explanations in (b,c) show relevant regions for the model's decision using light and irrelevant ones using dark colors.}}
    \label{fig:pascal_voc}
\end{figure}

Specifically, we applied our data augmentation strategy to a decoy variant of fashion-MNIST \cite{teso2019explanatory}, based on \cite{xiao2017fashion} (\textit{cf.} Methods).
The average test set accuracy of a multilayer-perceptron (with the same hyperparameters as in~\cite{ross2017right}) is reported in Tab.~\ref{tab:results}(a) for three correction strategies: no corrections, our \ce, and the input-gradient constraints (\rrr). The models' explanations for \ce are computed with \lime. Additionally, for every training image, we added $c = 1, 3, 5$ counterexamples where the decoy pixels are randomized. When no corrections are given, the accuracy on the test set is $48\%$: the confounders completely fool the network, \textit{cf.} Tab.~\ref{tab:results}(b). Providing even a single counterexample increases the accuracy to $82\%$, i.e., the effect of confounders drops drastically. With more counterexamples, the accuracy of \ce is similar to that of \rrr. Both methods pose valid improvements, thus showing that explanatory interactive learning (XIL) is an effective mean for correcting ``Clever Hans'' moments in machine learning and may even improve predictive performance and beliefs.

\rebuttal{Furthermore, we conducted experiments on the PASCAL VOC 2007 \cite{pascal-voc-2007} dataset. We focused on a five-class subset (\textit{cf.} Methods) and revised the model using XIL with the \rrr loss. Fig.~\ref{fig:pascal_voc} presents some example images and their explanations with and without user feedback, i.e. default (test accuracy.: $78\%$) and XIL trained (test accuracy: $73\%$). One can see that the classifier has learned the confounding factor for horse images (the source tag on the bottom left corner) without user feedback. After retraining the classifier using user feedback on the location of the source tag, we can see that the model no longer focuses on the confounder,
demonstrating the benefit and effectiveness of XIL also in this setting. Similar benefits can be observed on MSCOCO using \hint-like extensions. They may help to more quickly align human and gradient-based network explanations, as shown in the supplement.}

\paragraph{Deep plant phenotyping: High predictive performance.} 
Next, we showcase the extent, importance, and usability of XIL. To this end, we performed classification and revised corrections of the learned models on a real-world, scientific dataset. This dataset corresponds to RGB and hyperspectral (HS) (\textit{cf.} Methods) images of leaf tissue from inoculated (Cercospora beticola) and healthy sugar beet plants. Notably, there is a strong variability in the extent of disease severity over all samples, with some samples clearly showing the characteristic of Cercospora Leaf Spot (CLS) (two rightmost samples in Fig.~\ref{fig:story}) while others do not (second to the left sample in Fig.~\ref{fig:story}) and for the human eye appear indistinguishable---at least in RGB---from healthy leaves (top sample in Fig.~\ref{fig:story}). Roughly $50\%$ of inoculated tissue samples showed visible CLS.

We performed classification using convolutional neural networks (CNNs) on the RGB and HS datasets (\textit{cf.} Methods).
The task was to classify the leaf samples into the one of the \rebuttal{two} classes: healthy \rebuttal{or} diseased. The corresponding \rebuttal{average} balanced accuracies determined over 5 cross-validation runs are shown in the left column (no corr.) of Tab.~\ref{tab:results}(b). They show high accuracies of 88\% on the RGB dataset and nearly perfect performance of 99\% on the HS dataset. It seems the HS data contains more relevant information to the classification task.

\paragraph{Be careful! The deep network \rebuttal{might} be right for the wrong reasons.}
\begin{figure}[t!]
\centering
\includegraphics[width=1.\linewidth]{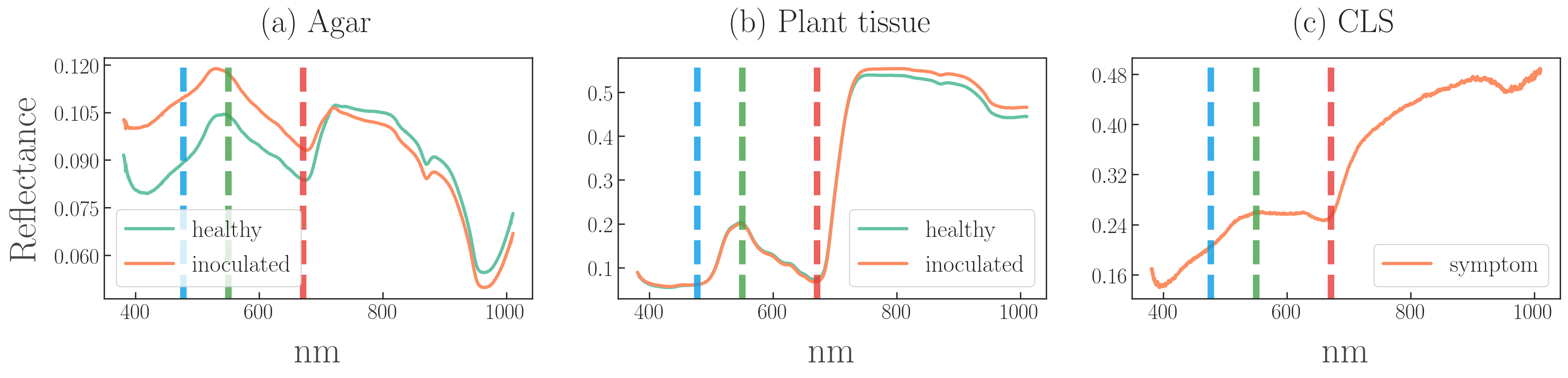}
\caption{Spectral signatures of measured agar plates with sugar beet leaf discs. Signatures were extracted of agar on which healthy and inoculated sugar beet leaf discs were placed (a), of healthy and inoculated sugar beet leaf discs (b) and \textit{C. beticola} symptoms of sugar beet leaves (c). Signatures were extracted from 100 pixels for each group and the mean value is presented. The vertical (green, blue, red) lines correspond to the wavelength selected for the pseudo-RGB images.}
\label{fig:leafspectra}
\end{figure}

\begin{figure}[t!]
\centering
\includegraphics[width=.8\linewidth]{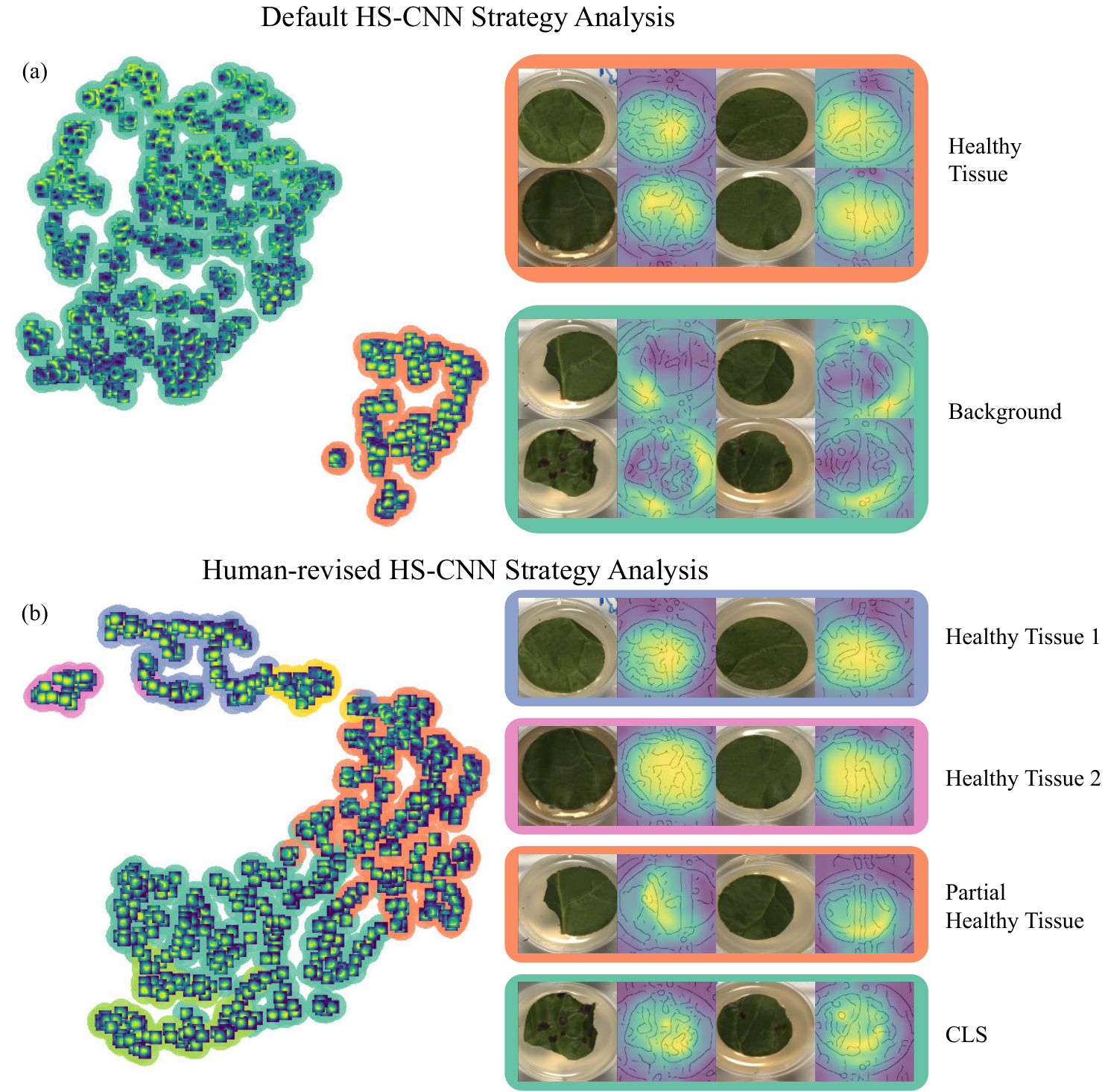}
\caption{\rebuttal{Cluster analysis of the different decision strategies after training CNNs on the HS data with the cross-entropy loss (Default) in (a) and with the \rrr loss in (b). The images are visualized in a two-dimensional t-SNE embedding and colored by the spectral clustering assignments.}}
\label{fig:strategieshs}
\end{figure}

The nearly perfect predictive performance is rather suspicious since plant phenotyping is a rather difficult task.
Therefore, we wanted to know the reasons for the predictions and visualized the explanations of the networks using \gradcams. 
\rebuttal{Specifically, we applied a spectral clustering and t-SNE \cite{maaten2008visualizing} analysis, similar to \cite{lapuschkin2019unmasking}, on the resulting explanations.}
\rebuttal{Fig.~\ref{fig:strategieshs}(a) shows the strategies of the CNN trained on the HS data for data points belonging to the test set only. Fig.~S.1(a) shows the strategies of the CNN trained with RGB data.} 
One can identify that the HS-CNN has altogether two prediction strategies, one for each predicted class label (\textit{cf.} supplement for more details). In the case of control samples, the HS-CNN focuses on large areas of the tissue, however, for inoculated samples, even if CLS are visible, the network focuses on the nutritional solution (agar) to classify these as inoculated. Moreover, when analyzing the reflectance of the agar across different stages of disease development, we could indeed identify differences between control and inoculated nutrition solution. This can be seen in the left panel of Fig.~\ref{fig:leafspectra}. Given the much smaller data dimensionality of the RGB images compared to the HS data, it seems likely that the RGB-CNN would have more difficulties focusing only on the agar as a classification feature, thus explaining the different classification strategies between HS and RGB-CNNs as well as the reduced classification performance of the RGB-CNN, compared to the HS-CNN.

\rebuttal{In any case}, both CNNs showed high to very high performances by largely using confounding factors within the dataset. The trained neural networks used strategies, which a biologist would consider as cheating rather than valid problem-solving behavior. The accuracies may not correspond to the true performance when measured in an environment outside of the lab setting, possibly even leading to dangerous consequences if left untackled.

\paragraph{\rebuttal{Revising} the model to be right for the right reasons.} 
It is too simple to say that we can not trust these models and even question if machines are truly ``intelligent''. We now show that with the human in the loop revising the machine, as in the XIL setting, the models can recover from the observed ``Clever Hans''-like strategies towards trustful ones.

To this end, we let a \rebuttal{plant biologist} revise the machine by constraining the machine's explanations to match her domain knowledge. 
Since the used models are differentiable, we focused on using \rrr rather than using the CE strategy, though both would be valid here within the XIL framework. 
%\rebuttal{Since, she was uncertain about what a valid explanation should be, we did not make use of \hint.}   
%Specifically, we added a second regularization term to our loss function such that we penalize the model during training when it uses an area that might exploit a dataset artifact. 
\rebuttal{Specifically, we simulated the interaction between the domain experts and the ML models. After training a model without any interactions, plant physiologists analyzed the provided predictions and corresponding explanations. She decided that it is always a wrong reason to focus on the background and consequently her annotations corresponded to binary masks of the whole tissue (\textit{cf.} Methods)}.

% \begin{figure}[ht!]
% \centering
% \includegraphics[width=.8\linewidth]{images/compressed/Strategy_analysis_rrr.pdf}
% \caption{Cluster analysis of the different decision strategies after training CNNs with a cross-entropy loss and a right for the right reason loss. The top row specifies the strategies of a CNN trained with RGB images. The bottom row specifies the strategies of a CNN trained with hyperspectral images. \rebuttal{The same figures as in Fig. \ref{fig:strategiesdefault} and Fig. \ref{fig:strategiesrrr} however divided by data type rather than training mode can be found in the supplementary material.}}
% \label{fig:strategiesrrr}
% \end{figure}
%

%\rebuttal{Similar to} the standard training mode %(i.e. cross-entropy loss and L2 weight regularization) 
As before, we analyzed the decision strategies of the \rrr trained model using t-SNE and spectral clustering. 
%\rebuttal{The results are summarized in Fig.~\ref{fig:strategiesrgb}(b) for the RGB-CNN and Fig.~\ref{fig:strategieshs}(b) for the HS-CNN.}
\rebuttal{The results are summarized in Fig.~\ref{fig:strategieshs}(b) for the HS-CNN and Fig.~S.1(b) for the RGB-CNN.}
%As one can see, the number of decision strategies has largely decreased after training the RGB-CNN with \rrr. 
% Similar to the default training mode %(i.e. cross-entropy loss and L2 weight regularization) 
% we analyzed the decision strategies of the \rrr training mode using t-SNE and spectral clustering. The results can be seen in Fig.~\ref{fig:strategiesrrr}, with the explanations of the RGB-CNN in (a), whereas those of the HS-CNN located in (b). \rebuttal{These results show explanations of data only in the test set.}%As one can see, the number of decision strategies has largely decreased after training the RGB-CNN with \rrr. 
%In comparison, after training the HS-CNN with \rrr, 
As one can see, after training the HS-CNN with \rrr, the model focuses on image regions lying only on the tissue, regardless of the underlying class. The strategies of control samples correspond to nearly full activation of the whole tissue, whereas for inoculated samples the identified relevant image regions are often \rebuttal{very specific spots}. Particularly, the model now focuses on the CLS, which were previously essentially ignored.
Fig. \ref{fig:story}(d) shows in more detail several examples of the observed strategies used by the corrected HS-CNN in comparison to the observed ``Clever Hans'' strategies of the unrevised machine.
Although the model's performance slightly decreased, \textit{cf.} Tab.~\ref{tab:results}(b), it is still able to classify samples without visible symptoms. 
%However, 
Even exploring different hyper-parameters for \rrr, we were not able to force the RGB-CNN to fully ignore the background,
%\rebuttal{as illustrated in Fig.~1(b). of the supplementary material}
\rebuttal{as illustrated in Fig.~S.1(b).}
% Fig.~\ref{fig:strategiesrrr} (a) showcases this. %The unchanged accuracy for the RGB-CNN with \rrr of Tab.~\ref{tab:results} must, therefore, be taken with a grain of salt. 
As shown in Fig.~\ref{fig:leafspectra}(a), 
although the HS-CNN has much more information at hand to focus on the confounding factors in the first place. However, after revision with \rrr, it is easier for the HS-CNN to make accurate predictions based on the reflectance of the tissue in comparison to the RGB-CNN (Fig.~\ref{fig:leafspectra}(b)). 
Particularly, the HS-CNN mainly uses a spectral area 
%---the third left heatmap of Fig.~\ref{fig:spatial_explanations}--- 
for prediction, which is beyond the RGB area. This explains the difficulty of correcting the RGB-CNN.

\rebuttal{We now focus on evaluating the default and revised models on a non-confounded test dataset to investigate the generalization improvement of training with XIL. Due to a missing non-confounded test set for the scientific dataset, we performed the simple trick of replacing the confounding features of all test samples with other values (\textit{cf.} Methods). The results are summarized in Tab. \ref{tab:results}(c), reporting the average test accuracy over five cross-validations. One can see that indeed the accuracy of the revised model is higher than that of the default model for both modifications.
These results further indicate the generalization improvements due to XIL. Further experiments applying prior knowledge can be found in the supplement.}

\paragraph{\bf Trust development during XIL.}
After demonstrating that explanations and especially XIL are necessary to reveal and correct so-called ``Clever Hans'' behavior of ML models we finally investigate how explanations influence the trust of users in the learning process.
To this end, we designed a questionnaire about a machine that learns a simple concept by querying labels (but \emph{not} explanation corrections) to an annotator. The online questionnaire was administered to 106 participants of varying ages and backgrounds.

Specifically, we designed a toy binary classification problem of ($3 \times 3$) black-and-white images, inspired by the color dataset used in ~\cite{ross2017right}. The subjects were told that an image is positive if the two top corners are white and negative otherwise. They were shown 30 images together with the classification of an AI model and a knowledgeable annotator. The learning of the model was simulated by increasing the model's classification accuracy from $50\%$ over $70\%$ to $100\%$ after every 10 images.
\begin{figure}[t!]
    \centering
    \includegraphics[width=1.\linewidth]{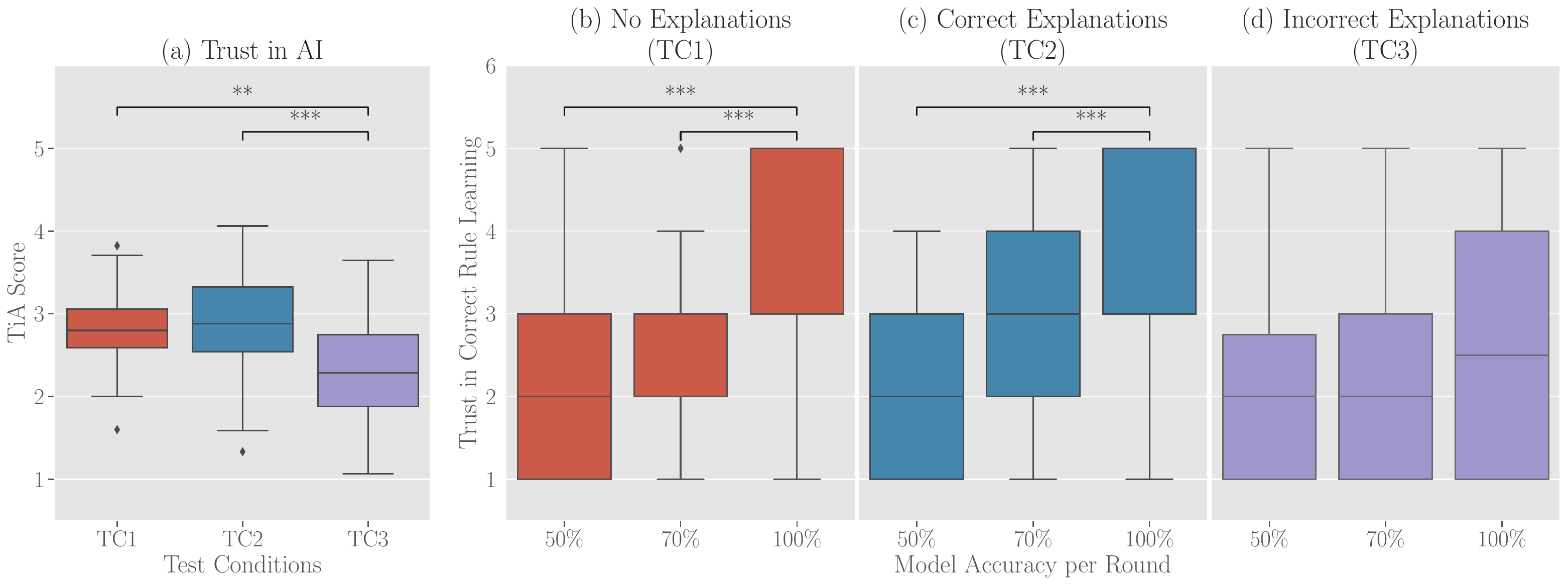}
    \caption{Results of the user study \rebuttal{on trust development}. (a) shows the total TiA Score over the three test conditions and (b-d) show in detail trust development (Q1) in correct rule learning after the three different learning stages of model accuracy (50\%, 75\%, 100\%) for each test condition. Only statistically significant results are highlighted. The center line of the box plots represents the median of the data, the box the interquartile distance between the first and third quartile, the whiskers the minimum and maximum value, discarding outliers which are plotted individually above the whiskers.
    The number of asterisks indicate the P values: $*$ $ P \leq 0.05$, $**$ $P\leq 0.01$, $***$ $P \leq 0.001$.
    \label{fig:user_study}}
\end{figure}
Each participant was randomly assigned to perform one of three experimental conditions with varying feedback from the model.
%In TC1 33 people participated, TC2 had 36 subjects, 38 people participated in TC3 and TC4 37
In test condition 1 (\textbf{TC1}), the participant received feedback for each image in the form of the model's prediction and the label provided by a knowledgeable annotator. No explanations were shown. Test conditions 2 and 3 (\textbf{TC2}, \textbf{TC3}) were identical to TC1, meaning that at every stage \emph{the same example, prediction and feedback label} were shown, but now explanations were also provided. The explanations highlighted the two most relevant pixels in form of red dots. In TC2, the explanations converged to the correct rule---they highlight the two top corners---from the \nth{6} image onwards. In TC3 the explanations converged to an incorrect rule---an image was classified as positive if the two top right pixels were white---from the \nth{12} image onward.
To assess the participant's trust in the model's skills we used the Trust in Automation Questionnaire (TiA) %by K\"orber 
\cite{korber2018theoretical}. After each learning process stage, the subjects were asked to rate (\textbf{Q1}) ``I trust that the AI has learned the correct rule for classifying such images.''. Lastly, having seen all images, subjects were asked to answer the full TiA. %Only subjects of TC2 and TC3 had to answer items of the explanation quality category after the TiA.

Fig. \ref{fig:user_study} summarizes the results, where (a) shows the total TiA score over TC1-TC3 and (b-d) the Q1 results for each test condition over the different stages of the learning process.
%As one can see, trust in AI varies between different variations of explanations. More precisely, the results indicate a slightly positive trust impact when providing correct explanations compared to no explanations. However, providing incorrect explanations results in a significant loss in the user's trust in the AI system. 
\rebuttal{ They confirm previous findings: without explanations, people trust highly accurate machines, but the trust drops when wrong behavior is witnessed~\cite{hoffman2013trust}. 
Users expect machines and their explanations to be correct. Indeed, explanations may increase the trust in earlier iterations at lower predictive performances, if they are correct. But people do not forgive wrong explanations if the predictions are correct. Thus, users really care about the ``right for the wrong reasons'' case.}

\rebuttal{
Taking all our empirical results together, people care about ``Clever Hans''-like moments in machine learning, XIL can eliminate them, and XIL may even improve the model’s predictive performance.
}

\section*{Conclusion}
In recent years, AI methods, especially machine learning with various directions and algorithms \cite{jordan2015machine, ghahramani15Probabilistic}, have become more and more successful in a wide range of areas like computer vision, natural language processing, and robotics, among others. Consider e.g. AlphaZero surpassing human-level performance in playing chess and Go. During its self-play training process, AlphaZero discovered a remarkable level of Go knowledge. This included not only fundamental elements of human Go knowledge, but also non-standard strategies beyond the scope of traditional human Go knowledge \cite{silverMastering2017}, exemplifying the potential of these methods to discover strategies previously unknown even to experts of the domain. 
However, studies from various applications such as  \cite{zech2018confounding,badgeley2019deep,chaibub2019permutation,lapuschkin2019unmasking} have revealed that learning machines can also result in ``Clever Hans''-like moments, i.e., human-undesired strategies where the machine exploits artifacts in the dataset.  

To ``un-Hans'' machines, we introduced the novel learning setting of ``explanatory interactive learning'' (XIL) and illustrated its benefits. XIL adds the scientist into the training loop. She interactively revises the original model via providing feedback on its explanations, used to automatically augment the training with counterexamples or to modify the model using \rrr.  \rebuttal{Our experimental results demonstrate that users care strongly about ``Clever Hans''-like moments in machine learning and XIL can indeed help avoiding them.} 
%encourages (or discourages) trust into the underlying model. \rebuttal{This is necessary to turn AI into an enabler of scientific discovery.}
%This shows that the vision of Donald E. Knuth---``{\it Let us change our traditional attitude to the construction of programs. Instead of imagining that our main task is to instruct a computer what to do, let us concentrate rather on explaining to human beings what we want a computer to do.}''---is not insurmountable for machine learning. 

There are several possible avenues for future work to overcome the current limitations of XIL.  
\rebuttal{
Acquiring annotations, especially of explanations, can be time consuming. The number of interactions required in order to reach an acceptable state is an open issue~\cite{teso2019explanatory}.}
\rebuttal{Hence, one should work on optimal query strategies for XIL that aim at minimizing the interaction efforts. Adapting regret bounds from co-active learning~\cite{shivaswamy2015coactive} might be an interesting alternative. 
Moreover, the data at hand may not always allow XIL to fully alleviate wrong reasons without decreasing the network's predictive performance. One should develop ways for keeping the drop as small as possible.}
\rebuttal{Furthermore, XIL relies on two assumptions, namely, (a) faithful explanations can be computed, 
and (b) the user feedback is faithful, too. Assumption (a) is still subject to very active research,
particularly for deep learning methods \cite{adebayo2018sanity} (see the supplement). One should improve the quality and robustness of XAI methods and also explore XIL for interpretable models~\cite{chenLTBRS19ThisLooksLikeThat}.
If the user is rather confident about the right reasons, learning to explain methods such as \hint provide an interesting avenue for future work. Our initial results, see the supplement, are encouraging. However, 
even scientific experts do not always know the reasons for predictions.
Therefore, one should strive to better understand the effects of wrong feedback and even adversarial attacks \cite{dombrowskiAAAMK19} on XIL.}
\rebuttal{Additionally, one should turn other interactive learning settings such as coactive~\cite{shivaswamy2015coactive}, active imitation~\cite{judah2012active}, mixed-initiative interactive~\cite{cakmak2011mixed} and guided probabilistic learning~\cite{odom2018human} into explanatory one. Lastly, because it is not yet clear what makes explanations good for humans \cite{narayanan2018humans}, one should extend explanatory interactions towards using alternative explanations, multiple modalities and counterfactuals \cite{kanehira2019learning, huk2018multimodal}. In any case, interacting with explanations of machine learning models is an enabler for scientific discoveries for humans and machines in cooperation.%of humans and machines together. 
%partnership of machines and humans and an enabler for scientific discovery.  % is not insurmountable. 
%\rebuttal{Additionally, it is valuable to extend the XIL framework to further methods, possibly incorporating different aspects of the framework, e.g. \cite{selvaraju2019taking}. A very relevant and interesting avenue for future work is ``co-actively learning to explain'' and making explanations robust against attacks \cite{dombrowskiAAAMK19} (see a detailed discussion of related work and limitations in Methods).} Finally, building inherently interpretable machine learning models \cite{chenLTBRS19ThisLooksLikeThat} is an exciting approach to understanding a model's decision. However, even in this setting, XIL should be beneficial and should help the user to understand and appropriately build trust in the model's decisions. Building inherently interpretable machine learning models \cite{chenLTBRS19ThisLooksLikeThat} is an exciting approach, however, even in this setting, XIL should be beneficial to help the user understand and appropriately build trust in the model's decisions. 
}

\section*{Methods}

\begin{figure}[tbh]
    \centering
    \includegraphics[width=.8\linewidth]{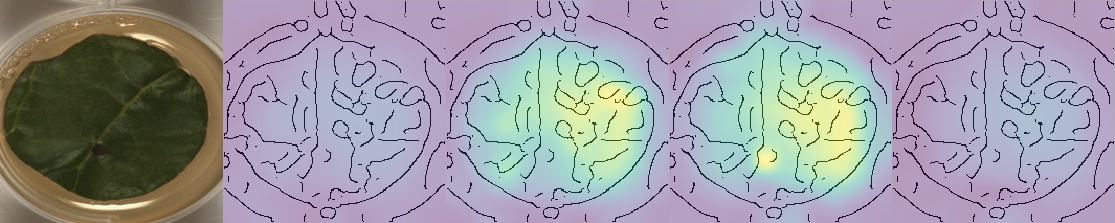}
    \caption{\gradcams \rebuttal{of a hyperspectral sample} with spatial and spectral explanations of a corrected network. Leftmost image shows the sample followed by the corresponding spatial activations maps mapped to four different hyperspectral areas. The areas are 380-537 nm, 538-695 nm, 696-853 nm and 854-1010 nm.\label{fig:spatial_explanations}}
\end{figure}

\paragraph{\bf Active learning.}  The active learning paradigm targets scenarios
where obtaining supervision has a non-negligible cost. Here we cover the
basics of pool-based active learning, and refer the reader to two excellent
surveys~\cite{settles2012active,hanneke2014theory} for more details.  Let
$\mathcal{X}$ be the space of instances and $\mathcal{Y}$ be the set of labels (e.g. $\mathcal{Y}
= \{\pm 1\}$).  Initially, the learner has access to a small set of labeled
examples $\mathcal{L} \subseteq \mathcal{X} \times \mathcal{Y}$ and a large pool of unlabeled
instances $\mathcal{U} \subseteq \mathcal{X}$. The learner is allowed to query the label
of unlabeled instances (by paying a certain cost) to a user functioning as an annotator, often a
human expert.  Once acquired, the labeled examples are added to $\mathcal{L}$ and
used to update the model.  The overall goal is to maximize the model quality
while keeping the number of queries or the total cost at a minimum.  To this
end, the query instances are chosen to be as informative as possible, typically
by maximizing some informativeness criterion, such as the expected model
improvement~\cite{roy2001toward} or practical approximations thereof.  By
carefully selecting the instances to be labeled, active learning can enjoy
much better sample complexity than passive
learning~\cite{castro2006upper,balcan2010true}.  Prototypical active learners
include max-margin~\cite{tong2001support} and Bayesian
approaches~\cite{krause2007nonmyopic}; recently, deep variants have
been proposed~\cite{gal2017deep}.
However, active (showing query data points) and even coactive learning (showing additionally the prediction of the query data point)
%is not enough to
do not establish trust: informative selection
strategies just pick instances where the model is uncertain and likely wrong.  There
is a trade-off between query informativeness and user
``satisfaction'', as noticed and explored in~\cite{schnabel2018short}.
To properly modulate trust into the model, we argue it is
essential to present explanations, e.g., visual ones as shown in Fig.~\ref{fig:spatial_explanations}.

\paragraph{\bf Local explainers.} There are two main strategies for explaining machine
learning models. Global approaches aim to explain the model by converting it
\emph{as a whole} to a more interpretable
format~\cite{Bucilua2006Model},\cite{bastani2017interpreting}.  Local explainers
instead focus on the arguably more approachable task of explaining
\emph{individual predictions}~\cite{lundberg2016unexpected}.  While explainable
interactive learning can accommodate any local explainer, in our implementations
we used either \lime~\cite{ribeiro2016should} or \gradcam~\cite{selvaraju2017grad}, both described next.

The idea of \lime (Local Interpretable Model-agnostic Explanations) is simple: even though a classifier may rely on many uninterpretable features, its decision surface around any given instance can be locally approximated by a simple, interpretable \emph{local model}.  In \lime, the local model is defined in terms of simple features encoding the presence or absence of \emph{basic components}, such as words in a document or objects in a picture. \rebuttal{While not all problems admit explanations in terms of elementary components, many of them do~\cite{ribeiro2016should}; in this case, \lime assumes these to be provided in advance.} An explanation can be readily extracted from such a model by reading off the contributions of the various components to the target prediction and translating them into an interpretable visual artifact. For instance, in document classification one may highlight the words that support (or contradict) the predicted class.

\gradcams are a generalization of Class Activation Maps, introduced by \cite{zhou2016learning} and take advantage of the facts that, firstly, deeper layers of a CNN capture higher-level visual constructs and, secondly, that convolutional features retain spatial information. As such, the last convolutional layer represents a trade-off between high visual representation and spatial information. Specifically, 
a \gradcam is computed by forward passing an image through the network, applying a backpropagation of a one-hot encoding vector that specifies the class label of interest up to the last convolutional layer. The resulting gradients of each channel are global average pooled, multiplied with the corresponding feature maps, summed and finally passed through a RELU activation function. In this way, the final feature maps of the convolutional feature extractor are weighted by the importance of these features. 
The resulting two-dimensional heatmap can finally be interpolated to the original input size for visualization. In case a 3D convolutional network is used to classify hyperspectral data the resulting heatmap is three dimensional also showing activations along the spectral dimension of the data, \textit{cf.} Fig.~\ref{fig:spatial_explanations}.

\begin{figure}[t!]
     \centering
     \includegraphics[width=0.8\linewidth]{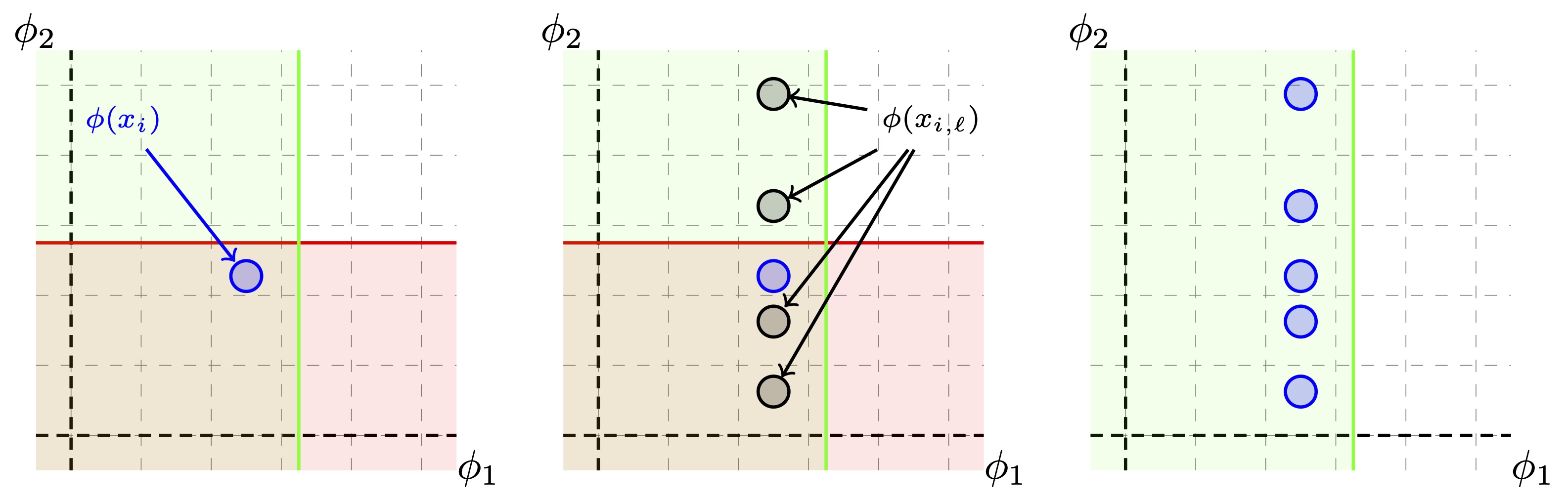}
     \caption{
         \label{fig:math}
         \rebuttal{Mathematical intuition for the counterexample strategy, exemplified for linear classifiers. Two data features are shown, $\phi_1$ and $\phi_2$, of which only the first is truly relevant. Left: The positive example $x_i$ is not enough to disambiguate between the red and green classifiers. Middle: Counterexamples $x_{i,\ell}$ are obtained by randomizing the irrelevant feature while keeping the label of $x_i$.  The counterexamples approximate a (local) orthogonality constraint. Right: The red classifier is inconsistent with the counterexamples and eliminated.  See the Methods section \emph{Explanatory Interactive Learning with counterexamples} for details.}
         (Best viewed in color).
     }
\end{figure}

\paragraph{\bf Explanatory Interactive Learning with counterexamples.}
Why is this data augmentation a sensible idea?
To see this, consider the case of linear max-margin classifiers. Let $f(x) = \inner{\vw}{\vphi(x)} + b$ be a linear classifier over two features, $\phi_1$ and $\phi_2$, of which only the first is relevant. Fig.~\ref{fig:math} shows that $f(x)$ (red line) uses $\phi_2$ to correctly classify a positive example $x_i$. In order to obtain a better model (e.g. the green line), the simplest solution would be to enforce an orthogonality constraint $\inner{\vw}{(0, 1)^\top} = 0$ during learning. Counterexamples follow the same principle. In the separable case, the counterexamples $\{\bar{x}_{i\ell}\}_{\ell=1}^c$ amount to additional max-margin constraints~\cite{cortes1995support} of the form $y_i \inner{\vw}{\vphi(\bar{x}_{i\ell})} \ge 1$. The only ones that influence the model are those on the margin, for which strict equality holds. For all pairs of such counterexamples $\ell, \ell'$ it holds that $\inner{\vw}{\vphi(\bar{x}_{i\ell})} = \inner{\vw}{\vphi(\bar{x}_{i\ell'})}$, or equivalently $\inner{\vw}{\vdelta_{i\ell} - \vdelta_{i\ell'}} = 0$, where $\vdelta_{i\ell} = \vphi(\bar{x}_{i\ell}) - \vphi(x_i)$.  In other words, the counterexamples encourage orthogonality between $\vw$ and the correction vectors $\vdelta_{i\ell} - \vdelta_{i\ell'}$, thus approximating the orthogonality constraint above.
Most importantly, this data augmentation procedure is model-agnostic, although alternatives indeed exist: (manually) adding a discovered data artifact to samples of
other classes~\cite{anders2019analyzing}, contrastive examples~\cite{zaidan2007using}, feature ranking~\cite{small2011constrained} for SVMs and constraints on the input gradients for differentiable models~\cite{ross2017right}.

\rebuttal{We note that due to sampling, \lime\ may output different explanations for the same prediction. To reduce the variance of the experiments with \ce of Tab. \ref{tab:results}, we ran it $10$ times and retained the $k$ components identified most often as relevant by \lime.}

\paragraph{fashion-MNIST dataset.}
The fashion-MNIST dataset, a fashion product recognition dataset%\footnote{\url{https://github.com/zalandoresearch/fashion-mnist}}
, includes 70,000 images over 10 classes. All images were corrupted by introducing confounders, that is, $4 \times 4$ patches of pixels in randomly chosen corners whose shade is a function of the label in the training set and random in the test set (see~\cite{ross2017right} for details).

\rebuttal{\paragraph{PASCAL VOC 2007 dataset.}
We used a subset of the PASCAL VOC 2007 dataset in our experiment. This subset includes resp. 1470 train and 782 test images over 5 classes (horse, cat, bird, bus, dog). Only samples from the \textit{horse} class contain confounding features, i.e. watermark text.
We rescale all the images to 224*224*3 to use the VGG-16 network \cite{simonyan2014very} as classifier, and we used the ImageNet-pre-trained weights as initial weights, as well as the ADAM optimizer \cite{KingmaB14}. We trained a default model without user feedback and a model with user feedback for 2k epochs. The explanation method was instantiated with input gradients (IG).}
 
\paragraph{Sample collection.} 
To demonstrate the significance of XIL, we demonstrate XIL for deep plant phenotyping and plant disease detection, a growing and relevant field of research~\cite{esther14naturereview,souza09nature,tardieu2017cell,pound2017deep,mochida2018computer, mahlein2019quantitative}.
To this end, we recorded a scientific, real-world dataset---a plant phenotyping dataset consisting of RGB and hyperspectral images (HS) of healthy and diseased sugar beet leaves. Then, we applied convolutional neural networks to classify the plants' leaves into the categories \textit{control} (healthy) and \textit{inoculated} (diseased) and investigated the underlying reasons for the network's predictions. As a model disease, Cercospora leaf spot (CLS) was used. This is caused by \textit{Cercospora beticola} and is the most destructive leaf disease of sugar beet with worldwide economic importance. 

The dataset used in this study corresponds to HS and RGB images of leaf discs of sugar beet cv. Isabella (KWS, Einbeck, Germany) inoculated with \textit{Cercospora beticola}. Sugar beet seeds were pre-grown in small pots and piqued after the primary leaves were fully developed. The seedlings were then transferred into plastic pots (diameter of 17 cm) on commercial substrate (Topfsubstrat 1.5, Balster Erdenwerk, GmbH, Sinntal-Altengronau, Germany) under greenhouse conditions and watered as necessary. After reaching growth stage 16 according to BBCH scale \cite{BBCHScale} the plants were inoculated with \textit{C. beticola} conidia which were collected from infested sugar beet leaves after incubation in a moist chamber for 48 hours. A spore suspension of $5\times10^5$ was sprayed onto leaves before the plants were transferred into plastic bags to achieve 100\% RH for 48 hours. For image acquisition leaf discs were stamped out with a cork borer of 2 cm diameter and placed on 10g/l pyhtoagar (Duchefa Biochemie B.V, Haarlem, Netherlands), containing 0.34 mM benzimidazole, 10 g sucrose and 3 mg kinetin. To observe different symptom classes sugar beet leaves of 9, 14 and 19 days after inoculation (dai) were used since the first symptoms appeared 9 dai. As a control group, 18 leaf discs of untreated sugar beet plants were measured as well and five technical replications with 6 discs each were used for each symptom group.

Each sample, both control and inoculated, was measured daily over five consecutive days such that a sample from 9 dai reappears four further times in the dataset as 10 to 13 dai. A few samples were discarded due to technical issues. The percentage of healthy leaves to unhealthy leaves was approximately $26\%$ to $74\%$, respectively. For image acquisition leaf discs on agar were placed on a linear stage at a distance of 53 cm to a Hyperspec VNIR E-series imaging sensor (Headwall Photonics, Bolton, MA, USA) in the range of 380 nm to 1010 nm. The VNIR sensor has a spectral resolution of 2-3 nm and a pixel pitch of 6.5 $\mu$m. The sensor was surrounded by eight lamps (Ushio Halogen Lamp J12V-150WA/80 (Marunouchi, Chiyoda-ku, Tokyo, Japan)) and the distance between lamps and leaves was 60 cm with a vertical orientation of 45°. Exposure times of 44 ms were used for the VNIR sensor.

The dataset consists of 2410 samples with 504 samples labeled as control and 1906 labeled as inoculated. Control samples were not re-used as inoculated samples. \rebuttal{The collected hyperspectral raw data size was around 4TB. After preprocessing the data by cutting out the leaf discs into hyperspectral cubes the data size is 140 GB. Since there is a lot of redundancy in the wavelength resolution, we further sub-sampled the depth of the data cubes resulting in a final data size of 32GB.}

\paragraph{Data preparation.} As mentioned above, each sample was imaged over five consecutive days such that each sample, though slightly differing from day to day, is represented up to 5 times within the full dataset. In this way, a sample from 9 dai would occur for 4 further days (10-13 dai). To prevent the models from memorizing the structure of the individual leaf samples and correlating this to the corresponding labels, a precaution was taken to exclusively contain all days of one sample either in the training or validation dataset.

\rebuttal{\paragraph{Removing confounders for the scientific dataset.}
It is essential to maintain the underlying assumption that the training and test data are drawn from the same distribution. If this is not the case, changes in accuracy might be due to artifacts of different data, rather than deficits of the model \cite{hookerekk19}. We applied two variations to the test samples of the HS dataset to remove the confounders: we set the background (everything but the plant tissue) (1) to the per-channel average of the non-tissue regions or (2) the per-channel average of the full images of the training data. We then evaluated the default and \rrr revised CNNs on this modified test dataset. We focused here only on the HS data and model, due to the limitations of the RGB model's performance. }

\paragraph{RGB/HS classification.} The RGB images used for training the classifiers were generated from the hyperspectral data, by slicing the data at the corresponding RGB channels that were provided by the camera system (\textit{cf.} Fig.~\ref{fig:story} (A-Right)). Before training the RGB classifiers, the data was standard scaled following $z = (x - u) / s$, where $u$ is the mean and $s$ the standard deviation of the training samples. 

To train a classifier on the RGB images of sugar beet leaves we used a VGG-16 \cite{simonyan2014very} network pre-trained on ImageNet \cite{imagenet_cvpr09} to finetune the network parameters using the RGB plant images. For training a batch size of 32, a learning rate of 1e-4 and a step learning rate scheduler set to reduce the learning rate at epochs 5 and 15 by a factor of $0.1$ were used. Furthermore, the ADAM optimizer was used with L2 regularization 1e-5. Five separate cross-validation folds were trained until convergence, using a data split of $0.75$ for training and $0.25$ for testing. Convergence was reached after 30 epochs.

To classify the HS data we trained a convolutional neural network (CNN) architecture with batch normalization using 3D convolution filters, rather than standard 2D filters, learning features not only along the image dimensions but also over the spectral dimensions. The used network is build up with four residual blocks, each containing one to three convolutional layers. The last two layers are fully connected layers with a final softmax activation function. The other layers use ReLU activations. During training the networks we used dropout to prevent overfitting. The network's parameters are trained with a stochastic gradient descent optimizer with momentum using a batch size of 10 HS images, a learning rate of 1e-4 and an L2 regularization of 1e-5. 

Five separate cross-validation folds were trained until convergence, using a data split of $0.75$ for training and $0.25$ for testing. Convergence was reached after 100 epochs.

\paragraph{Analyzing classification strategies of the model.}
%We expected the networks to use obvious symptoms of the plant disease such as the characteristic CLS typically occurring at later stages of the disease, but also non-visible symptoms for early stages of disease progression that are hidden in the spectral domain. Examples from varying stages of the disease progression are shown in Fig.~\ref{fig:story} (A) together with feature importance maps (saliency maps) in the bottom row that human experts consider as right reasons. Specifically, the leftmost example shows a healthy tissue sample. The second to the left example shows an inoculated sample at an early stage of disease progression, where no obvious disease characteristics are visible and even the human expert is unsure of the right reason for classifying such a sample. The two rightmost samples show inoculated samples displaying the characteristic disease spots with varying extent. Training a deep neural network to classify the two categories resulted in surprisingly high performance (B). Unmasking the reasons for the classifications using an approach similar to that of Lapuschkin \textit{et al.}~\cite{lapuschkin2019unmasking}, different reasons of the model can be identified, showing that the model is right for the wrong reasons, focusing incorrectly on areas outside of the tissue (C). Using XIL, however, a human user can correct this behavior (D). The corrected model produced accurate predictions for the right reasons.

Based on the results of \cite{adebayo2018sanity}, in which the authors performed sanity checks over a variety of saliency methods, we chose to investigate our model's explanations using Gradient-weighted Class Activation Mapping (\gradcam)\cite{selvaraju2017grad}.

To analyze the resulting strategies produced by the layer-wise relevance propagation method (LRP), the authors of \cite{lapuschkin2019unmasking} revert to using spectral clustering on the resulting heatmaps in a pipeline they termed 'SpRAy'. \rebuttal{This clustering served to receive an overview of the extent of the model's decision strategies.} 
We apply SpRAy in a similar way, however, rather than using the raw \gradcam heatmaps, we perform a discrete Fourier transformation on these beforehand to better differentiate different strategies \rebuttal{which we had previously identified from single samples}. In detail, the pipeline is as follows
\begin{itemize}
  \item Perform a discrete Fourier transform on downsized \gradcam heatmaps.
  \item Using the Euclidean distance for the RGB data and the Cityblock distance for the HS data compute a k-nearest neighbor graph of the Fourier transformed heatmaps, represented as an adjacency matrix, $C$.
  \item Compute the affinity matrix as suggested in \cite{von2007tutorial} as $A = max(C, C^T)$.
  \item Perform an eigengap analysis \cite{von2007tutorial} to estimate the number of clusters, k, within the dataset.
  \item Perform spectral clustering on the affinity matrix, given k from the previous step
  \item Perform a t-SNE analysis \cite{maaten2008visualizing} on the similarity matrix, estimated from the affinity matrix as in \cite{lapuschkin2019unmasking} as $S = \frac{1}{A + \epsilon}$, whereby $\epsilon \in [0, 1]$, here we used $\epsilon = 0.05$.
\end{itemize}

\paragraph{\bf Applying XIL to CNNs for scientific dataset.}
We produced the matrix $A$ (Eq. \ref{eqn:rrr}) corresponding to full tissue masks for each sample. Specifically, for each sample, we created a binary mask having values of zero within the tissue and values of one everywhere else, i.e. the background. In this way during training the gradients everywhere but on the tissue are to be minimized.

The network models were retrained from the same initial values as in the default training mode (using only the cross-entropy loss), however, now using \rrr. To choose the optimal $\lambda_1$ value, the resulting explanations were visually assessed. The five cross-validation folds of HS-CNN were thus trained until convergence between 200 and 280 epochs using a $\lambda_1 = 20$ value, with all other hyperparameters as in the default training mode. For training the RGB-CNN with \rrr the learning rate was reduced to a constant learning rate of 5e-05. Although applying a range of $\lambda_1$ values from $0.1$ to $1000$, using the RGB-CNN, no satisfactory convergence state could be reached in which the regularized model showed acceptable explanations for each cross-validation run. The accuracies in Tab.~\ref{tab:results} and the strategies presented in \rebuttal{Fig.~\ref{fig:strategieshs}(b) and Fig.~S.1(b)} 
%Fig.~\ref{fig:strategiesrrr} 
correspond to \gradcams of training the five cross-validation folds with $\lambda_1=1$ for up to 200 epochs.

\paragraph{\bf Extended related work.}
\rebuttal{%We have shown that users are indeed able to revise machine learning models interactively in real-world applications. 
Using XIL with \ce or \rrr, users either introduce counterexamples into the dataset and thus teach the learner not to depend on the irrelevant components or directly penalize the learner as soon as it uses irrelevant components, respectively. One important advantage of XIL is that the user does not have to be certain about the right reasons and instead can explore the learned reasons of the machine, in contrast to other procedures such as preprocessing the training set. }

\rebuttal{Recently, Selvaraju \textit{et al.}~\cite{selvaraju2019taking} presented a framework (\hint) similar to \rrr but instead of penalizing the wrong reasons it advises the network to use a specific visual area (right reasons). As \ce and \rrr, the \hint method could be embedded within the introduced XIL framework in case the users are certain about the right reasons. However, in many scientific applications such as the presented plant phenotyping dataset users are uncertain about what a valid explanation should be. In this case, removing wrong reasons might be preferable to applying right reasons.} 

\rebuttal{The possibility of bi-directional exchange between user and model due to interaction \cite{abdel2020and} also distinguishes XIL from approaches for feature selection such as feature masking and approaches that embed prior knowledge into the training process, e.g. \cite{erion2019learning}. Lastly, interactions also allow that the user can provide incomplete explanations, in other words: only if it is actually required, the user can revise incorrect aspects of a model's explanation.}

\rebuttal{
Finally, we present the XIL framework here for visual tasks and visual explanations only. With our definition of XIL, it is also applicable to other data domains like natural language processing, see e.g.~\cite{ross2017right, teso2019explanatory}. However, we experienced that explanations, i.e. right and wrong reasons, are more difficult to define for this modality. In future work one should generally address the best ways to present explanations, even in multi-modal scenarios.}

\paragraph{\bf Details on participant recruitment and study procedure.}
The presented study is part of an extensive thesis work \cite{herbert2019}.
It was conducted as an online survey, the link of which was distributed via the social network Facebook and the forum of the student body of the department of computer science at TU Darmstadt. Due to the distribution on these channels a wide range of people with different ages and different backgrounds was generated. Each participant completed only one of the three test conditions with 33 participants in TC1, 36 participants in TC2 and 37 participants in TC3, totaling 106 participants overall.

The wording of the original TiA was modified by replacing ``system'' with ``artificial intelligence (AI)''.The response format to each question was a 5-point rating scale from strongly disagree to strongly agree.

\paragraph{\bf Statistical analysis of the user study.}

%A missing value analysis was conducted in order to validate that the pairwise case exclusion could be conducted for the analysis. Missing values can either occur when subjects do not answer an item or when subjects choose the response option no response. Missing values should be missing randomly and not systematically, which was analyzed via Little’s MCAR (Missing Completely At Random) analysis. For randomly missing values the pairwise case exclusion is suitable. Missing values were also analyzed on item and subject level, in order to exclude items and subjects with many missing values.
Samples with missing values were removed from the analysis and for all tests a significance level with alpha being 5\% was used. 

For all tests with the same sample/samples, the alpha level was corrected via the Bonferroni-Holm method. The corrected alpha level will be stated for every analysis. For testing the hypotheses one multi-factorial analysis of variances (MANOVA) and several one-factorial ANOVAs were conducted. The ANOVA, as well as the MANOVA, requires normal distribution of data, independence of data as well as homogeneity of the variances. To test the latter a Levene-Test was conducted before every ANOVA and the MANOVA. Normal distribution was presumed due to the sample sizes and as the samples were drawn randomly the independence of data was also presumed. A significant result of an ANOVA / MANOVA means that at least two of the groups differ significantly with respect to the dependent variable, but it is not stated which groups differ. Therefore, if the carried out analyses of variances were significant, post hoc tests were carried out to investigate which groups differed exactly. Post hoc tests were selected in this study as the hypotheses did not point out which groups should differ, which is why every possible comparison had to be considered. For post hoc testing, the Tukey-HSD-Test and the Pairwise-Test were performed.

The TiA score of subjects being familiar with AI over the whole sample (all test conditions combined) was higher ($M = 2.82$, $SD = .64$) than the TiA score of subjects being unfamiliar with AI ($M = 2.51$, $SD = .59$). As the conducted Levene-Test ($F(5, 99) = 1.8$, $p = .12$, $\alpha = .05$) was not significant, the homogeneity of variance assumption held. Therefore, the MANOVA was conducted with a significant result for the independent variable test condition ($F(2 , 99 ) = 10.10$, $p < .001$, $\alpha = .025$). The MANOVA was significant for the independent variable familiarity with AI ($F (1 ,99 ) = 7.12$, $p = .009$, $\alpha = .025$). It was not significant for the interaction of the two independent variables ($F( 2, 99) = .28$, $p = .75$, $\alpha = .025$). 

For Fig.~\ref{fig:user_study}(a) in order to determine which test conditions differed significantly in their TiA score, a pairwise test was conducted as a post hoc test. The pairwise test showed significant differences between TC1 and TC3 ($p = .0016$, $\alpha = .05$) as well as between TC2 and TC3 ($p = .0003$, $\alpha = .05$).

For Fig.~\ref{fig:user_study}(b) the conducted Levene-Test was not significant ($F (2, 96) = .59$, $p = .56$, $\alpha = .05$). Therefore, an ANOVA was conducted afterwards and showed a significant result ($F (2, 96) = 33.83$, $p < .001$, $\alpha = .0125$). Trust in the correct rule learning by the AI was significantly different between the blocks. The conducted Tukey-HSD test found a significant difference in trust into the correct rule learning only between stage 1 and 3 ($p < .001$, $\alpha = .05$) and between stage 2 and 3 ($p < .001$, $\alpha = .05$).

For Fig.~\ref{fig:user_study}(c) the Levene-Test was not significant ($F (2, 104) = .28$, $p = .75$, $\alpha = .05$). The ANOVA was significant ($F (2, 104) = 23.19$, $p < .001$, $\alpha = .0167$). Therefore, a Tukey-HDS test was performed to investigate which blocks differed significantly. The test found only stage 1 and 3 ($p < .001$, $\alpha = .05$) and stage 2 and 3 ($p < .001$, $\alpha = .05$) to differ significantly with respect to trust in correct rule learning by the AI. 

For Fig.~\ref{fig:user_study}(d) the conducted Levene-Test was not significant ($F (2, 105) = 1.32$, $p = .27$, $\alpha = .05$). The afterwards conducted ANOVA was also not significant ($F (2, 105) = 1.62$, $p = .20$, $\alpha = .05$). Therefore, there was no significant difference in trust into correct rule learning by the AI in TC3 and no post hoc test was performed.

\section*{Data availability}
The ML benchmark Fashion-MNIST is available at
\url{https://github.com/zalandoresearch/fashion-mnist}. \rebuttal{The PASCAL VOC2007 dataset is available at \url{http://host.robots.ox.ac.uk/pascal/VOC/voc2007/}.}
The RGB and hyperspectral data that support the findings of this study are available at \url{https://tudatalib.ulb.tu-darmstadt.de/handle/tudatalib/2278.4} and in the code repository \url{https://codeocean.com/capsule/4559958/tree}.
The user study is available at \url{https://github.com/ml-research/xil/tree/master/Trust_Study}.

\section*{Code availability}
\rebuttal{The code and a fully runnable capsule to reproduce the figures and results of this article, including pre-trained models, can be found at \url{https://codeocean.com/capsule/4559958/tree}.}
%Code supporting this study is published online at \url{https://github.com/ml-research/xil}. The Python implementation of CAIPI for interactive explanatory learning is available at \url{https://github.com/stefanoteso/caipi}.

\section*{Statement of ethical compliance}
The authors confirm to have complied with all relevant ethical regulations, according to the Ethics Commission of the TU Darmstadt (\url{https://www.intern.tu-darmstadt.de/gremien/ethikkommisson/auftrag/auftrag.en.jsp}). An informed consent was obtained for each participant prior to commencing the user study.

\section*{Acknowledgments}
ST an KK thank Antonio Vergari, Andrea Passerini, Samuel Kolb, Jessa Bekker, Xiaoting Shao, and Paolo Morettin for very useful feedback on the conference version of this article. Furthermore, the authors are thankful to Frank J{\"a}kel for support and supervision on the user study, to Cigdem Turan for providing the figure sketches, and to Ulrike Steiner and Stefan Paulus for very useful feedback. PS, AKM, AB and KK acknowledge the support by BMEL funds of the German Federal Ministry of Food and Agriculture (BMEL) based on a decision of the Parliament of the Federal Republic of Germany via the Federal Office for Agriculture and Food (BLE) under the innovation support program, project ``DePhenSe'' (FKZ 2818204715). WS an KK were also supported by BMEL/BLE funds under the innovation support program, project ``AuDiSens'' (FKZ 28151NA187). ST acknowledges the supported by the European Research Council
(ERC) under the European Union’s Horizon 2020 research and innovation
programme, grant agreement No.~[694980] ``SYNTH: Synthesising Inductive Data
Models''. XS and KK also acknowledges the support by the German Science Foundation project
``CAML'' (KE1686/3-1) as part of the SPP 1999 (RATIO).
AKM was partially funded by the Deutsche Forschungsgemeinschaft (DFG, German Research Foundation) under Germany’s Excellence Strategy - EXC 2070 -- 390732324

\section*{Conflict of interest statement}
The authors declare the following competing interests: HS is employed by LemnaTec GmbH.

\section*{Author information}
{\bf Affiliations} \\ \\
Technical University of Darmstadt, Computer Science Department, Artificial Intelligence and Machine Learning Lab, Darmstadt, Germany\\
Patrick Schramowski, Wolfgang Stammer, Franziska Herbert, Xiaoting Shao\\ \\
Technical University of Darmstadt, Computer Science Department and Centre for Cognitive Science, Darmstadt, Germany \\
Kristian Kersting\\ \\
University of Trento, Department of Information Engineering and Computer Science, Trento, Italy \\
Stefano Teso\\ \\
University of Bonn, Institute of Crop Science and Resource Conservation (INRES) -- Plant Diseases and Plant Protection, Bonn, Germany \\
Anna Brugger\\ \\
Institute of Sugar Beet Research, Goettingen, Germany \\
Anne-Katrin Mahlein\\ \\
LemnaTec GmbH, Aachen, Germany \\
Hans-Georg Luigs\\ \\

\noindent
{\bf Author Contributions} \\ \\
PS and WS contributed equally to the work.
PS, WS, ST, KK designed the study. ST, KK designed and published (AAAI /ACM Conference on Artificial Intelligence, Ethics, and Society 2019) the preliminary version of this manuscript. PS, WS, XS, ST, and KK developed extensions of the basic XIL methods. PS, WS, AB, AKM, and KK interpreted the data and drafted the manuscript. AB and PS designed the phenotyping dataset. AB and HGL carried out the phenotyping dataset measuring. PS, WS, AB did the biological analysis. FH performed and analyzed the user study. AKM and KK directed the research and gave initial input. All authors read and approved the final manuscript. \\ \\
{\bf Corresponding author}\\ \\
Correspondence to Patrick Schramowski and Wolfgang Stammer.

\bibliography{scibib}

\end{document}